\documentclass[twoside,11pt]{article}

\usepackage{jmlr2e}
\usepackage{array}
\usepackage{url}
\usepackage{latexsym}
\usepackage[usenames,dvipsnames]{color}
\usepackage{amsmath}
\usepackage{amsfonts} 
\usepackage{enumitem}
\usepackage[caption=false]{subfig}
\usepackage{graphicx}
\usepackage[utf8]{inputenc}
\usepackage{multirow}
\ShortHeadings{Quantifying Mental Health from Social Media with Neural User Embeddings}{Amir, Coppersmith, Carvalho, Silva and Wallace}
\firstpageno{1}

\begin{document}

\title{Quantifying Mental Health from Social Media with\\ Neural User Embeddings}

\author{}
\author{\name Silvio Amir \email samir@inesc-id.pt \\
       \addr INESC-ID Lisboa, Instituto Superior Técnico, Universidade de Lisboa\\
       Lisboa, Portugal       
       \AND
       \name Glen Coppersmith  \email glen@qntfy.com \\
       \addr Qntfy \\
      Washington DC, United States
      \AND       
      \name Paula Carvalho \email pcc@inesc-id.pt \\
      \addr INESC-ID Lisboa, and Universidade Europeia, LIU\\
      Lisboa, Portugal
      \AND              
      \name M\'{a}rio J. Silva \email mjs@inesc-id.pt \\
      \addr INESC-ID Lisboa, Instituto Superior Técnico, Universidade de Lisboa\\
      Lisboa, Portugal
      \AND
      \name Byron C. Wallace \email b.wallace@northeastern.edu \\
      \addr Northeastern University\\
      Boston MA, United States}

\maketitle

\begin{abstract}
%Yet preventing and diagnosing mental health problems is difficult because it requires patients to recognize the associated symptoms and actively seek help. Moreover, 
  Mental illnesses adversely affect a significant proportion of the population worldwide.  However, the methods traditionally used for estimating and characterizing the prevalence of mental health conditions are time-consuming and expensive. Consequently, best-available estimates concerning the prevalence of mental health conditions are often years out of date. Automated approaches to supplement these survey methods with broad, aggregated information derived from social media content provides a potential means for near real-time estimates at scale. These may, in turn, provide grist for supporting, evaluating and iteratively improving upon public health programs and interventions. 
  %At the same time, web-based social networks provide a comfortable medium for people to anonymously connect with others affected by similar issues. This provides an opportunity to develop automated tools for identifying at-risk individuals early on, thus affording the opportunity for effective prevention and treatment. In aggregate, such automated monitoring at scale may allow more accurate estimates concerning the extent of population-level mental illness incidence. 
  % bcw 4/23 -- would be interested in hearing Glen's take, but I find the epidemiological view more convincing here rather than diagnosing/detecting individuals
 
 We propose a novel model for automated mental health status quantification that incorporates \emph{user embeddings}. This builds upon recent work exploring \emph{representation learning} methods that induce embeddings by leveraging social media post histories. Such embeddings capture latent characteristics of individuals (e.g., political leanings) and encode a soft notion of homophily. In this paper, we investigate whether user embeddings learned from twitter post histories encode information that correlates with mental health statuses. To this end, we estimated user embeddings for a set of users known to be affected by depression and post-traumatic stress disorder (PTSD), and for a set of demographically matched `control' users. We then evaluated these embeddings with respect to: (i) their ability to capture homophilic relations with respect to mental health status; and (ii) the performance of downstream mental health prediction models based on these features. Our experimental results demonstrate that the user embeddings capture similarities between users with respect to mental conditions, and are predictive of mental health.

\end{abstract}
    
\section{Introduction}

Mental illness is a critically important concern, significantly and adversely affecting a wide swath of the population directly and indirectly. An estimate by the Centers for Disease Control from 2008~\citep{centers2010current}, suggests that 9\% of US adults may meet the criteria for depression at any given time. While not as prevalent as depression, post traumatic stress disorder (PTSD) issues still cost hundreds of billions of dollars worldwide, according to a conservative estimate from the NIH\footnote{\url{https://www.nimh.nih.gov/health/statistics/cost/index.shtml}}.
The collective effect of mental health conditions, as measured by Daily Adjusted Life Years (DALYs), exceeds that of malaria, war, or violence\footnote{For a visualization of DALYs see \url{https://vizhub.healthdata.org/gbd-compare/}}~\citep{whiteford2013global}. At the same time, mental health problems are often difficult to identify and thus treat. For example, perhaps half of depressive cases go undetected, in part due to the heterogeneous and complex expression of this condition~\citep{paykel1997defeat}. Another exacerbating factor is that diagnosis generally requires individuals to actively seek out treatment. Yet, the manifestation of this condition and prevailing social stigmas may disincline afflicted individuals to seek treatment. 

The internet may provide a comfortable medium for people to express their feelings anonymously and connect with health-care professionals~\citep{mccaughey2014best} and others affected by similar conditions~\citep{de2016discovering}. Furthermore, individuals openly discuss mental health challenges on public social network platforms such as Twitter \citep{coppersmith_acl_2014,coppersmith_cross_condition_2015}. Prior work has demonstrated the potential of using social media to investigate mental health issues~\citep{paul2011you}, including depression~\citep{schwartz2014towards}, PTSD~\citep{coppersmith_icwsm_2014} and suicidal ideation \citep{coppersmith_suicide_clpsych_2016,de2016discovering} in individuals. However, models and techniques to identify and quantify mental health related signals from social media are relatively novel. Interest in these applications has motivated the creation of a shared task for the Computational Linguistics and Clinical Psychology workshop (CLPsych)\footnote{\url{http://clpsych.org}}, which aimed to advance the state-of-the-art in technologies capable of discriminating users affected by mental illness from controls, given their post history~\citep{shared_task_2015_overview}. A variety of methods have been proposed for this task, but none have achieved consistently superior performance, which implies that improvements may yet be realized by improved models.

Neural representation learning methods have been shown capable of automatically discover good representations (e.g., predictive features) from data, freeing practitioners from the burden of manually designing and encoding task-specific features~\citep{Bengio-et-al-2015-Book,goldberg2016primer}. In natural language applications, this has resulted in neural distributed representations becoming the \emph{de-facto} standard representational approach. \emph{Word embeddings} in particular aim to implicitly encode latent word semantics (in the distributional sense), and can be learned in an unsupervised fashion by means of predictive models that exploit word co-occurrence statistics and other regularities in unlabeled corpora~\citep{bengio2003neural}. These models have been recently extended to infer representations for larger textual units~\citep{le2014distributed}, and even user representations~\citep{li2015learning,amir2016modelling}. It has been shown that these \emph{user embeddings} also capture latent user aspects and can be used in downstream applications, such as sarcasm detection~\citep{amir2016modelling} and content recommendation~\citep{yu2016user}. 

In this paper we investigate whether user representations induced via neural models can inform clinical models operating over social media. In particular we consider whether user embeddings (learned directly from historical data) capture aspects of mental health status. To this end we leverage the dataset created for the CLPsych shared task to address two research questions: (1) To what extent do user embeddings capture information relevant for mental health analysis applications, over social media? (2) Can user embeddings be leveraged to discriminate between users suffering from mental illness and demographically matched controls? We answer the first question by comparing different approaches to induce user representations from a collection of posts. In particular, we investigated whether the induced embeddings capture homophilic relations between users with respect to mental health. To answer the second question, we developed and evaluated predictive models, leveraging user embeddings, to discriminate users affected by depression, PTSD, and age- and gender-match controls. 

The main contributions of this paper are as follows: (i) we show that unsupervised user embeddings induced from posting histories capture user similarities, and are predictive of mental health conditions; (ii) we develop a novel neural model that incorporates and refines these embeddings to improve the categorization of users with respect to mental health status; furthermore, we show that the resultant fine-tuned user embeddings better align with mental health conditions.

%\begin{enumerate}
%    \item \emph{To what extent do user embeddings capture information relevant for mental health analysis applications, over social media?}
%    \item \emph{Can user embeddings be leveraged to discriminate between users suffering from mental illness and demographically matched controls?}
%\end{enumerate}

The remainder of the paper is organized as follows. The next section introduces the aforementioned CLPsych shared task and the corresponding dataset. Section \ref{sec:related} reviews the literature on user modelling for social media analysis and neural embedding learning. In Section \ref{sec:model}, we formally describe the user embedding model used in our experiments, and discusses the connections with prior representation learning methods. Section \ref{sec:embedding_analysis} addresses the first research question by evaluating the properties captured by the user embeddings. Section \ref{sec:prediction} reports on the classification experiments we conducted to answer the second research question. Finally, we present our conclusions in Section \ref{sec:conclusions}.

%%\vspace{
\section{Depression and PTSD on Twitter}
\label{sec:task}
%%\vspace{-.5em}
In 2015, the CLPsych workshop held a shared task to foster progress in NLP technologies with potential for applications related to mental health analysis, over social media streams~\citep{clpsych_workshop_2015,shared_task_2015_overview}. To that end, a dataset was compiled comprising users that have publicly stated on Twitter that they were diagnosed with depression (327 users) or 
%post-traumatic stress disorder
PTSD (246 users), and an equal number of randomly selected demographically-matched users as \emph{controls}\footnote{This data was collected according to the ethical protocol of~\cite{benton2017ethical}, and follows the recommendations spelled out in~\cite{mikal2016ethical}.}. For each user in this dataset, associated metadata and posting history was also collected --- up to the 3000 most recent tweets, per limitations of the Twitter API. For more details on the construction and validation of the data, see \citep{shared_task_2015_overview,coppersmith_acl_2014,coppersmith_icwsm_2014}. 

The participants were then asked to develop models to discriminate between users affected by mental illness from controls, given their posts and metadata. Specifically this entailed three binary sub-tasks: (i) \texttt{depression vs control}, (ii) \texttt{PTSD vs control} and (iii) \texttt{depression vs PTSD}. The proposed systems were based on a wide range of approaches including: rule-based systems leveraging lexical decision lists~\citep{pedersen2015screening}, linear classifiers exploiting features based on word clusters and topic models~\citep{preotiuc2015mental}, supervised topic models~\citep{resnik2015beyond} and systems exploiting character-level language models~\citep{shared_task_2015_overview}. %For more details on these methods, we refer to the corresponding papers.
However, we note that none of the proposed systems performed consistently better than the others across all the sub-tasks and evaluation metrics, highlighting the difficulty of this problem. Moreover, none of these systems used explicit representations of \emph{users}, which is the innovation we propose here. We did not participate in this shared task, and thus could not obtain the official test data. Therefore, our results are not directly comparable to those of the participating teams. Nevertheless, we compared our proposed approach with the majority of the previously proposed methods.

%Furthermore, the participants were separately evaluated in the three sub-tasks. 

%For real-world applications, however, we believe that a multiclass classification approach would be preferable in practice. % It seems unreasonable to force a model to decide if someone suffers from depression or PTSD when most likely he or she is affected by neither (given the prevalence of these illnesses).  % bcw 4/23 -- i do not understand the last sentence. what are we trying to say here? i think something like PTSD and depression should practically be grouped for our purposes, which is reasonable, but this is not clear at all in the text right now.

%%\vspace{-.5em}
\section{Related Work}
\label{sec:related}
%%\vspace{-.5em}

Most of the research in social media analysis has been concerned with deriving better models that operate on representations of the texts comprising individual users posts, both via manually crafted features and, more recently, representation learning approaches~\citep{severyn2015twitter,astudillo-EtAl:2015:ACL-IJCNLP}. Nevertheless, for a variety of problems it is crucial to also capture characteristics of the \emph{users} involved in the communications. These include, information extraction~\citep{yang2016toward}, opinion mining~\citep{tang2015learning}, sarcasm detection~\citep{bamman2015contextualized} and content recommendation~\citep{yu2016user}. The most straightforward approach to induce user representations is by scrapping ``profile'' information from social websites, to manually extract features based on social ties, demographic attributes, or posting habits~\citep{bamman2015contextualized,rajadesingan2015sarcasm}. However, these approaches require significant effort for data collection, and for task- and domain-specific feature engineering. Furthermore, the available user profile information depends on specific social websites and may not always be available. It may also be inaccurate or simply outdated. 

%\vspace{-.75em}
\subsection*{Neural Embedding Learning}
%\vspace{-.65em}

In recent years, models in NLP have moved from \emph{discrete} word representations, based on scalars representing indices into a pre-defined vocabulary, towards \emph{distributed} continuous vector word representations that encode latent semantics --- these are usually referred as \emph{word embeddings} \citep{goldberg2016primer}. The general framework to learn unsupervised word embeddings involves associating words with parameter vectors, which are then optimized to be good predictors of other words that occur in the same contexts~\citep{bengio2003neural}. \textsc{Skip-Gram}~\citep{mikolov2013distributed}, one of the most popular word embedding models, operationalizes this approach by sliding a \emph{window} of a pre-specified size across the corpus. At each step, the center word is used to predict the probability of one of the surrounding words, sampled proportionally to the distance to the center word. \cite{le2014distributed} later expanded this approach with two \textsc{Paragraph2Vec} models that also learn representations for paragraphs (or, more broadly, any sequence of words): (i) \textsc{PV-dm}, tries to predict the center word of the sliding window, given the surrounding words \textbf{and} the paragraph (i.e., their respective embeddings); and (ii) \textsc{PV-dbow}, tries to predict the words of a sliding window within a paragraph, conditioned only on the respective paragraph embedding.

Recently proposed methods to learn user representations use essentially the same approach --- associate users with parameter vectors, and optimize these to accurately predict observable attributes or the words used by said user in previous posts~\citep{li2015learning,amir2016modelling,yu2016user}. As discussed above, leveraging user profile information to collect attributes is not always reliable. Interestingly, however, the embeddings induced by \cite{amir2016modelling}, using only the previous posts from a user, were shown to capture latent user aspects (e.g. political leanings) and a soft notion of `homophily' --- \emph{similar} users were generally represented with similar vectors. Furthermore, these user representations were successfully used to improve a downstream model for sarcasm detection in tweets. Similarly, \cite{yu2016user} used user embeddings to improve a microblog recommendation system. Our work aims to ascertain if user embeddings learnt only from previous posts, can capture useful signals for clinical applications.

%\vspace{-.75em}
\section{Learning User Embeddings}
%\vspace{-.75em}
\label{sec:model}
To learn user embeddings, we adopted an approach similar to that recently proposed by \cite{amir2016modelling}. The idea is to capture relations between users and the content (i.e., the words) they generate, by optimizing the probability of sentences conditioned on their authors. Formally, let $\mathcal{U}$ be a set of users, $\mathcal{C}_j$ be a collection of posts authored by user $u_j \in \mathcal{U}$, and ${S=\{w_1, \ldots, w_N\}}$ be a post composed of words $w_i$ from a vocabulary $\mathcal{V}$. The goal is to estimate the parameters of a user vector $\mathbf{u}_j$, that maximize the conditional probability:
\begin{equation}
\begin{split}
     P(\mathcal{C}_j|u_j) & \propto \sum_{S \in \mathcal{C}_j} \sum_{w_i \in S} \log 
     P(w_i|\mathbf{u}_j) 
\end{split}
\label{eq:user_embeddings}
\end{equation}
However, directly estimating these quantities (e.g., with a log-linear model) would require calculating a normalizing constant over a potentially large number of words, a computationally expensive operation. Because we are only interested in the user vectors $\mathbf{u}_j$ and not the actual probabilities as such, we can approximate the term $P(w_i|\mathbf{u}_j)$ by minimizing the following Hinge-loss objective: 
\begin{flalign}
    \mathcal{L}(w_i, u_j) = \sum_{\tilde{w}_k \in \mathcal{V}}\max(0,1 - \mathbf{w}_i \cdot \mathbf{u}_j + \tilde{\mathbf{w}_k} \cdot \mathbf{u}_j) 
\end{flalign}
\noindent where word $\tilde{w}_k$ (and associated embedding, $\tilde{\mathbf{w}}_k$) is a \emph{negative sample}, i.e. a word not occurring in the sentence under consideration, which was written by user $u_j$. By learning to discriminate between observed positive examples and \emph{pseudo}-negative examples, the model shifts probability mass to more plausible observations~\citep{smith2005contrastive}. Note that we represent both \emph{words} and \emph{users} via $d$-dimensional embeddings --- word embeddings, $\mathbf{w}_i \in \mathbb{R}^d$ which are assumed to have been pre-trained trough some neural language model; and user embeddings $\mathbf{u}_j \in \mathbb{R}^d$ to be learned. We will refer to this approach as \textsc{User2Vec}.\footnote{This formulation is a simplification of~\cite{amir2016modelling} model. Specifically, we omitted a term in Eq.\ref{eq:user_embeddings}, encoding the marginal probability of $S$; and we allow the negative samples to be drawn from all the words in $\mathcal{V}$. These simplifications dramatically reduce training time without significant loss of quality on the resulting embeddings.}

We note that barring some minor operational differences, this model is equivalent to the \textsc{PV-dbow} variant of \textsc{Paragraph2vec} --- if users are viewed as paragraphs. The key differences are that: (i) \textsc{User2Vec} predicts \textbf{all} the words in a post, whereas \textsc{PV-dbow} slides a window along the paragraph and only predicts one word per step; and (ii) \textsc{User2Vec} assumes that the word embeddings are pre-trained, whereas \textsc{PV-dbow} aims to jointly learn the word and paragraph vectors. 

%%\vspace{-.5em}
\section{User Embedding Analysis}
%%\vspace{-.5em}
\label{sec:embedding_analysis}

\begin{figure*}[!t]
  \centering
    \includegraphics[width=1\linewidth]{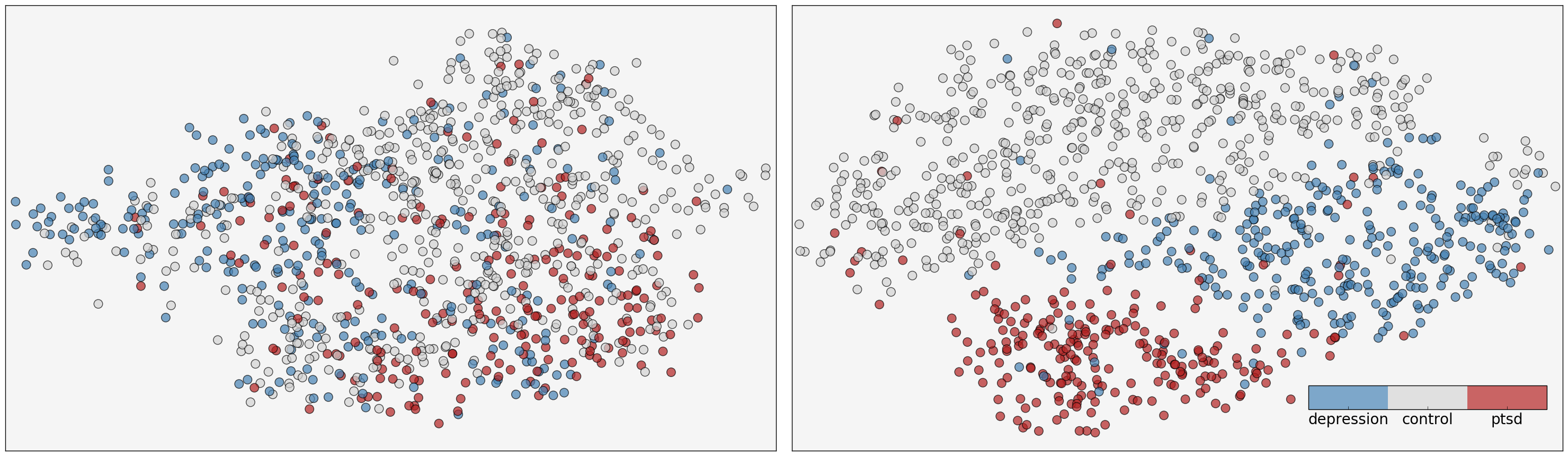}
    \caption{\label{fig:tsne} \textsc{PV-dm} embeddings projected into two dimensions, and colored according to the respective cohort. The plot on the left-hand side shows the original unsupervised embeddings --- despite being trained without labels, in this space, users tend to be surrounded by others from the same cohort. The plot on the right-hand side, shows the result of an embedding subspace projection induced with the NLSE model --- we can see that, in this adapted space, the users are better clustered by cohort.}
\end{figure*}

In this section we address our first research question by investigating whether user embeddings learned directly from social media data encode information relevant for public health applications.  Previous work has shown that these representations capture latent user aspects and a soft notion of `homophily'. If some of these aspects correlate with mental health, then the embeddings could be used to identify risk groups; for example, we might identify and characterize users that `look' like patients affected by depression. The ability to do so would potentially enable scalable, real-time estimates concerning the prevalence of mental health issues in particular populations.%, might warrant a subsequent investigation or intervention.

We estimated \textsc{User2Vec} embeddings from the shared task dataset described in Section \ref{sec:task}, as follows.\footnote{Code will be made available after publication} First, we pre-processed tweets by: lower-casing; reducing character repetitions to at most three repetitions; and replacing usernames and URLs with a canonical form. Users with fewer than 100 tweets were discarded. Next, we pre-trained a set of \textsc{Skip-Gram} word vectors from the task data and another large unlabeled Twitter corpus, using the \texttt{Gensim}\footnote{\url{http://radimrehurek.com/gensim/}} python package~\citep{rehurek_lrec}. Finally, for each user $u_j \in \mathcal{U}$, we sampled a held-out set $\mathcal{H}_j \subset \mathcal{C}_j$ with 10\% of the posting history. The rest of the data was used to estimate an embedding $\mathbf{u}_j$ by minimizing Eq. \ref{eq:user_embeddings} via stochastic gradient descent, using $P(\mathcal{H}_j|\mathbf{u}_j)$ as early stopping criteria. 

%\begin{enumerate}
%\item We pre-processed tweets by: lower-casing; reducing character repetitions to at most three repetitions; and replacing usernames and URLs with a canonical form. Users with fewer than 100 tweets were discarded.

%%\vspace{.35em}
%\noindent 2. 
%\item We pre-trained a set of \textsc{Skip-Gram} word vectors from the task data and another large unlabeled Twitter corpus, using the \texttt{Gensim}\footnote{\url{http://radimrehurek.com/gensim/}} python package~\citep{rehurek_lrec}.

%%\vspace{.35em}
%\noindent 3. For each user $u_j \in \mathcal{U}$, we sampled a held-out set $\mathcal{H}_j \subset \mathcal{C}_j$ with 10\% of the posting history. The rest of the data was used to estimate an embedding $\mathbf{u}_j$ by minimizing Eq. \ref{eq:user_embeddings} via stochastic gradient descent, using $P(\mathcal{H}_j|\mathbf{u}_j)$ as early stopping criteria. 

% bcw 4/24 -- "10% of the posting history"; is this split up temporally? i.e., 10% of data from the 'future' or are we training on tweets "from the future"? A reviewer may ask. The way I read this, though, is that this is just a set for early stopping anyway, so really not so important assuming that's right!
%\item For each user $u_j \in \mathcal{U}$, we sampled a held-out set $\mathcal{H}_j \subset \mathcal{C}_j$ with 10\% of the posting history. The rest of the data was used to estimate an embedding $\mathbf{u}_j$ by minimizing Eq. \ref{eq:user_embeddings} via stochastic gradient descent, using $P(\mathcal{H}_j|\mathbf{u}_j)$ as early stopping criteria. 
%\end{enumerate}
%%\vspace{.25em}
The same dataset was used to derive embeddings with \textsc{PV-dm} and \textsc{PV-dbow} models (also via \texttt{Gensim}). To ensure a fair comparison, we used the same hyper-parameters for all the models, which were set as follows: window size ${w=5}$, negative sample size ${s=20}$ and vector size ${d=400}$.

%%\vspace{-.5em}
\subsection{Measuring Homophily}
%%\vspace{

\begin{figure}[t!]
\subfloat[User vector similarity ranking. The first row corresponds to a 'query' user, and the columns show the top 100 most similar users, colored according to their class.]{%
  \includegraphics[width=1\columnwidth]{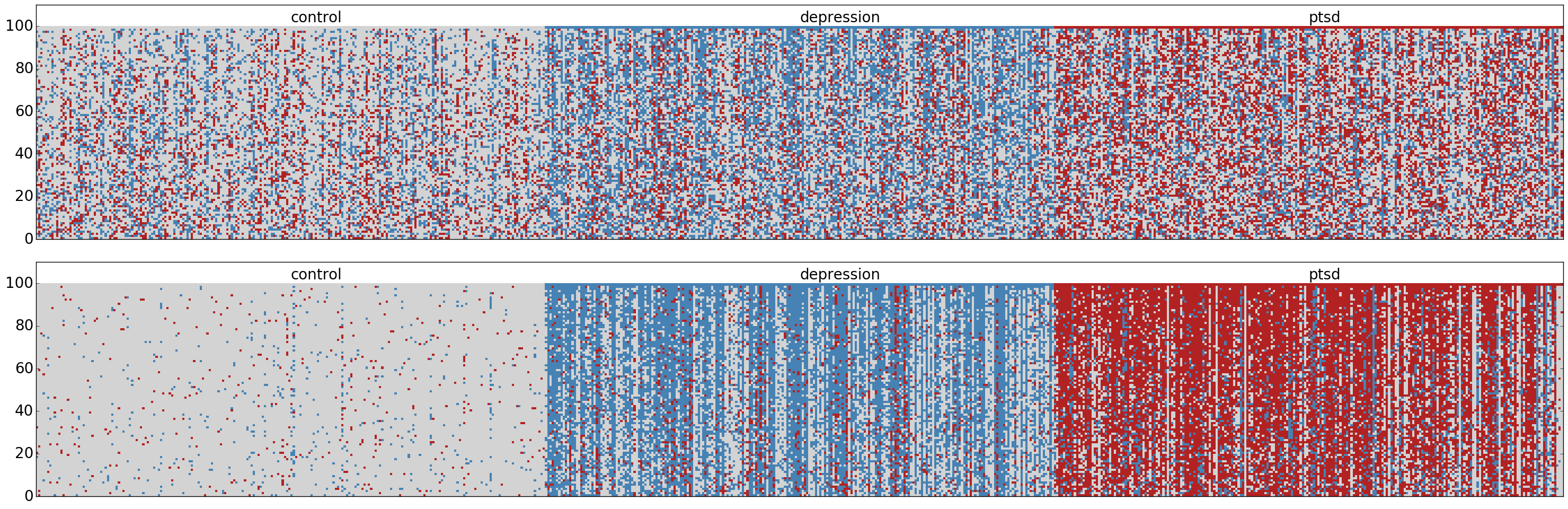}
  \label{fig:user_rank}
}
~\\
\subfloat[ROC curves and AUC scores of the induced user similarity rankings, per class.]{%
 \centering

    \includegraphics[width=1\columnwidth]{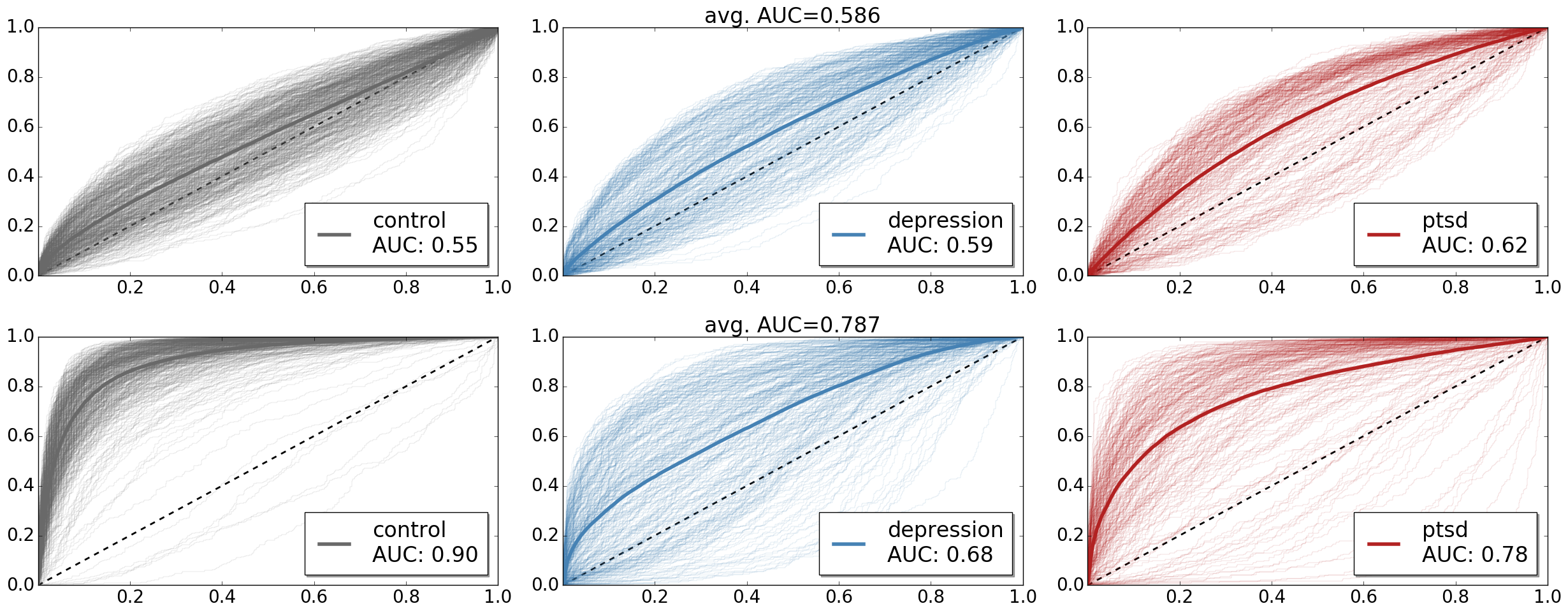}  
    \label{fig:rank_roc}

}

\caption{Measuring homophilic relations with respect to mental conditions with vector distances over the user embedding space. The top-most sub-plots refer to rankings induced with the \textsc{PV-dm} model, and the ones at the bottom correspond to rankings obtained with embeddings adapted with the NLSE model.}
\label{fig:homophily}
\end{figure}

To investigate if the induced user vectors capture homophilic relations with respect to mental health status, we first projected the $d$-dimensional vectors into a 2-dimensional space using t-Stochastic Neighborhood Embedding (TSNE)~\citep{van2008visualizing}. In Figure \ref{fig:tsne}, we plot the resulting points colored according to the respective class. One can see that, at least to some extent, embeddings do seem to capture some notion of homophily, i.e. users are often surrounded by others of the same cohort. 

To better quantify this effect and allow comparison of different user embedding models, we proceeded as follows. For each user in the corpus, we calculated the similarities to all \emph{other} users (i.e., the cosine similarity between their respective embeddings), inducing a ranking of users in terms of similarity to the `query' user. Intuitively, we would hope to see that users in the same mental health categories are comparatively similar to one another, i.e., users suffering from depression are most similar to other users also afflicted with depression. 

Figure \ref{fig:homophily} shows the results obtained with \textsc{PV-dm} vectors. The induced similarity ranking is shown in Figure \ref{fig:user_rank}, where the first row correspond to the query users, and each column shows their top $k=100$ most similar users, colored according to their class. Figure \ref{fig:rank_roc} shows the respective Receiver Operating Characteristic (ROC) curves under the induced ranking. In general, we found that all the user embedding models are able to capture user similarities. The embeddings induced with the \textsc{User2Vec}, \textsc{pv-dbow} and \textsc{pv-dm} all perform significantly better than chance, with AUC scores of 0.57, 0.57 and 0.59, respectively. Detailed plots can be found in the Appendix. The fact that user vectors are more likely to be close to those of others in the same cohort demonstrates that this approach does indeed capture signals relevant to mental health. This aligns with prior work showing that one's choice of words can be indicative of psychological states and mental health~\citep{pennebaker2001linguistic}. 

%Figure \ref{fig:homophily} shows the results obtained with \textsc{user2vec} vectors. The induced similarity ranking is shown in Figure \ref{fig:user_rank}, where the first row correspond to the query users, and each column shows their top $k=100$ most similar users, colored according to their class. Figure \ref{fig:rank_roc} shows the respective Receiver Operating Characteristic (ROC) curves under the induced ranking. Surprisingly, we found that all models have roughly the same ability to capture user similarities. While realizing different distributions of AUC across classes, all models attained the same average AUC of 0.55. Detailed plots can be found in the Appendix. The fact that user vectors are more likely to be close to those of others in the same cohort demonstrate that this approach does indeed capture signals relevant to mental health. This aligns with prior work showing that one's choice of words can be indicative of psychological states and mental health~\citep{pennebaker2001linguistic}. 

\section{Predicting Mental Health from Twitter Data}
\label{sec:prediction}

To address our second research question, we evaluated the user representations with respect to their predictive performance in downstream mental health analysis applications. Above we showed that representations induced from user posting histories encode relevant signals about mental health. But generic features estimated from unsupervised tasks are sub-optimal for downstream tasks~\citep{astudillo-EtAl:2015:ACL-IJCNLP}. Neural networks, when trained end-to-end, can refine generic embeddings to specific tasks by modifying their parameters during supervised training~\citep{collobert2011natural}. However, this strategy requires updating a large number of parameters, which is difficult when only small training datasets are available. The CLPysch dataset comprises 1094 labeled instances (after discarding users with fewer than 100 tweets). Given the modest size of the dataset, we adopted the 
\cite{astudillo-EtAl:2015:ACL-IJCNLP} Non-Linear Subspace Embedding approach, (NLSE), which is able to adapt generic representations to specific tasks with scarce labeled data.

\subsection{Proposed Model}

The NLSE model adapts generic embeddings to specific applications by learning linear projections into lower-dimensional subspaces, while the keeping the original embeddings fixed. Hence, the resulting \emph{embedding subspaces} capture domain- and task-specific aspects, while preserving the rich information encoded by the original embeddings. More formally, given an user embedding matrix $\mathbf{U} \in \mathbb{R}^{d\times|\mathcal{U}|}$, in which column $\mathbf{U}_{[j]}$ represents user $u_j \in \mathcal{U}$, we induce new representations by factorizing the input as $\mathbf{S} \cdot \mathbf{U}$ where $\mathbf{S} \in \mathbb{R}^{s \times d}$,  with $s \ll d$, is a (learned) linear projection matrix. Crucially, this imposes dimensionality reduction on the feature space, simultaneously eliminating noise and reducing the number of free parameters, making the model easier to train with small datasets. 

This model is similar to a feed-forward neural network with a word embedding layer and a single hidden layer. The main differences are: (1) the factorization of the embedding layer into two components (the original embedding matrix and a linear projection matrix) and, (2) the dimensionality reduction induced by the subspace projection, with typical reductions greater than an order of magnitude. Similar to feed-forward networks, we can use the backpropagation algorithm ~\citep{rumelhart1988learning} to jointly learn task-specific embeddings and the parameters for the classification layer. Using this approach, our proposed mental health prediction model can be formalized as: 
\begin{equation}
\begin{split}
     P(\mathcal{Y}| u_j) &\propto \boldsymbol{\beta} \cdot g(u_j) \\
     g(u_j) &= \sigma\left(\mathbf{S}\cdot \mathbf{U}_{[j]} \right)
\end{split}
\label{eq:nlse}
\end{equation}
\noindent where, $\sigma(\cdot)$ denotes an element-wise sigmoid non-linearity, and the matrix $\boldsymbol{\beta} \in \mathbb{R}^{|\mathcal{Y}| \times s}$ maps the embedding subspace to the classification space. Notice that at inference time this model reduces to a linear classifier with embedding subspace features.

% bcw 4/24 --you might want to define F1 somewhere just in case? remember this is a bit less commong outside of core NLP world
\subsection{Experimental Setup}

We evaluated embeddings induced with the \textsc{User2Vec} (\textsc{u2v}), and \textsc{Paragraph2vec}'s \textsc{PV-dm} and \textsc{PV-dbow} models (Section \ref{sec:embedding_analysis}), as features in Logistic Regression (LR) and \textsc{NLSE} classifiers. These were compared against baselines using textual features based on: 
% bcw 4/24 -- if we need space, suggest reducing spacing in enumerate -- find a stackoverflow thread to do this :)
\begin{enumerate} 
  \item \textsc{bow}: bag-of-words vectors with binary weights, $\mathbf{x} \in \{0,1\}^{|\mathcal{V}|}$; 
  \item \textsc{boe}: bag-of-embeddings. We leveraged \textsc{Skip-Gram} embeddings to build vectors, $\mathbf{x} = \sum_{w}\mathbf{E}_{[w]}$, where $\mathbf{E}_{[w]} \in \mathbb{R}^d$ is the embedding of word $w$;
  \item \textsc{lda}: bag-of-topics. We induced $t=100$ topics using Latent Dirichlet Allocation~\citep{blei2003latent}, to build vectors $\mathbf{x} \in \{0,1\}^t$ indicating the topics present in user's posts; 
  \item \textsc{bwc}: bag-of-word-clusters. We induced $k=1000$ ~\cite{brown1992class} word clusters, to build vectors $\mathbf{x} \in \{0,1\}^k$ mapping words in a user's posts to their respective clusters;
\end{enumerate}
We also evaluated baselines that combine user vectors with textual features (\textsc{u2v+bow} and \textsc{u2v+boe}).

% bcw 4/24 -- hmm but wait were splits w.r.t. *users* or *tweets*??? i presume the former
Experiments were conducted with a \mbox{10-fold} cross-validation protocol; at each iteration, the training partition was split into 80\% for model training and the remainder for validation purposes (i.e. hyper-parameter selection and early-stopping). For consistency, we used the same splits for all the models. We performed a grid-search to choose the best $\ell_2$ regularization coefficient, over the range $c=\{0.001,0.01,0.5,1,10,100\}$, for the LR models; and the optimal subspace size $s=\{10,15,20,25\}$ and learning rate $\alpha=\{0.01,0.1,0.5,1\}$, for the NLSE model. 

% bcw 4/24 -- can we make any claims about 'state of the artness' of results here? how are we faring compared to shared task participants??
\subsection{Results}
\label{section:results}
The classifiers were mainly evaluated with respect to the macro average $F_1$. We also report results in terms of \emph{binary} $F_1$, where we only average the scores for the \texttt{depression} and \texttt{ptsd} classes, to better ascertain the ability of the models to discriminate between mentally afflicted patients, which are less prevalent than the controls, but are the cases that we mostly care about.

\begin{figure*}[hbt]
  \centering
    \includegraphics[width=1\linewidth]{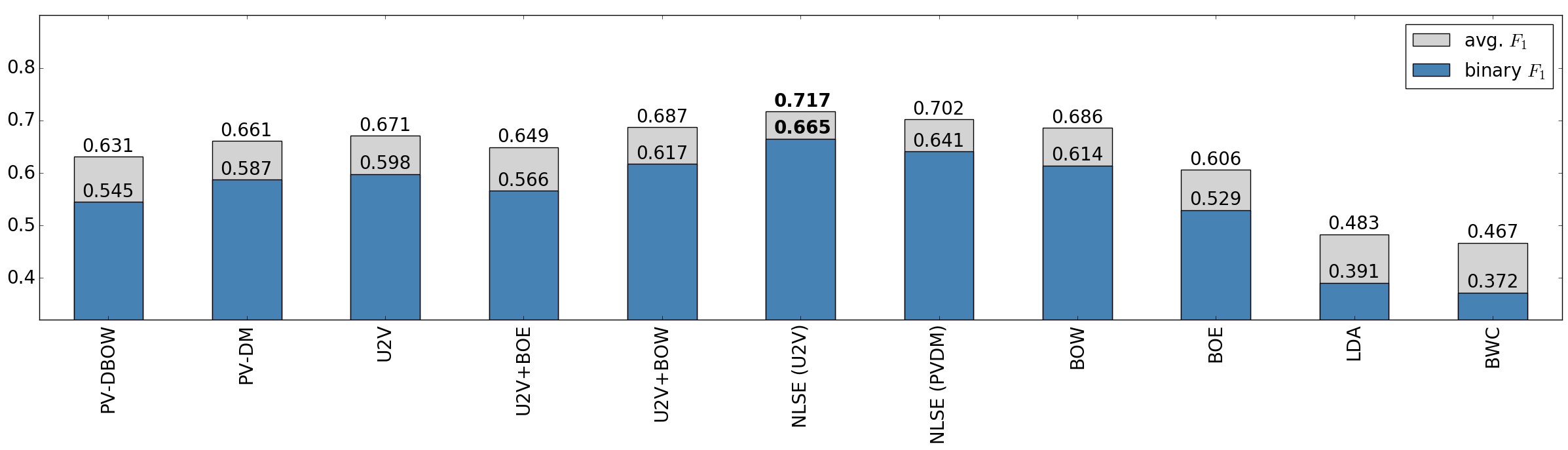}
    \caption{\label{fig:res} Performance of different models at discriminating users with respect to mental condition, in terms of $F1$ and \emph{binary} $F_1$.}
\end{figure*}

The main classification results are shown in Figure \ref{fig:res}. The first thing to note is that the \textsc{BOW} is a very strong baseline, essentially outperforming all the other linear classifiers based on textual features and user embeddings. One reason is that users affected with mental illnesses, often talk about their conditions and the \textsc{bow} model can easily pick-up on such clues. Regarding the user embeddings, we found that, despite being equivalent, the \textsc{PV-dbow} performed much worse than the \textsc{User2Vec}, showing that better embeddings can be obtained by trying to predict \textbf{all} the words in users posts, and leveraging pre-trained word vectors. On the other hand, the \textsc{PV-dm} model has a performance comparable to that of the \textsc{User2Vec}. 

As discussed above, generic embeddings are sub-optimal for downstream tasks, and our results are in line with this observation. By inducing task-specific representations via subspace projection, we were able to outperform all the other baselines by a fair margin. Note also that the NLSE approach is particularly better at discriminating the minority classes, i.e. patients suffering from \texttt{depression} and \texttt{ptsd}, as evidenced by the greater improvements in \emph{binary} $F_1$, when compared to the other baselines. To better understand the effects of the embedding adaptation, we repeated the same analysis described in Section 5 over the adapted embeddings (Figures \ref{fig:tsne} and \ref{fig:homophily}). We can see that the new representations are much better at discriminating the controls from the other users, suggesting that induced embedding subspace captures more fine-grained signal related to mental statuses.

\section{Conclusions}
\label{sec:conclusions}

In this paper, we investigated if embeddings induced from user posting histories capture relevant signals for clinical applications. In particular, we compared different user embedding methods, with respect to their ability to capture homophilic relations between users, and their performance as features in downstream mental health prediction models. The evaluation conducted over a dataset comprising of users diagnosed with depression and PTSD, and demographically matched controls, showed that these representations can indeed capture mental health related signals. This is in agreement with prior results from the field of psychology, establishing connections between word usage and mental status~\citep{pennebaker2001linguistic}. Interestingly, embeddings induced without knowledge of user labels capture similarities with respect to mental condition. Furthermore, we have shown that these embeddings can be tailored --- with a small amount of task-specific labeled data --- to capture more granular information, thus improving the quality of downstream models and applications. Ultimately, this work is a step toward more accurate inference concerning the mental health status of social media users, in turn enabling more accurate epidemiological real-time population-wide monitoring of mental health. Such accurate monitoring -- which is currently impossible -- may provide empirical support for increased resource allocation to programs dedicated to preventing and alleviating mental health issues. 

Moving forward, user embeddings may provide a pivotal piece to allow clinical psychologists to take full advantage of digital phenotyping data. In particular, learned user embeddings may provide a representation at a sweet spot between instantaneous (proximal) state and lifelong (distal) state, which is critical to understanding psychological phenomena and risk of crisis.

%Currently published user representations are largely ad-hoc and do not represent changes in a principled manner. User embeddings may provide a representation at the sweet spot between instantaneous (proximal) state and lifelong (distal) state, that is most critical to understandlying psychological phenomena and risk of crisis.

%\samir{@glen: maybe a punchline for the health-care people? @Silvio see the following and adjust as needed.}

%\byron{Suggest emphasizing potential uses of this -- real-time population monitoring of mental health, which could make the case for additional funding, etc.}

%\pagebreak

% ACKNOWLEDGEMENTS ONLY GO IN THE CAMERA-READY, NOT THE SUBMISSION
% \acks{Many thanks to all collaborators and funders!}

\bibliography{sample}

\begin{thebibliography}{40}
\providecommand{\natexlab}[1]{#1}
\providecommand{\url}[1]{\texttt{#1}}
\expandafter\ifx\csname urlstyle\endcsname\relax
  \providecommand{\doi}[1]{doi: #1}\else
  \providecommand{\doi}{doi: \begingroup \urlstyle{rm}\Url}\fi

\bibitem[Amir et~al.(2016)Amir, Wallace, Lyu, and Silva]{amir2016modelling}
Silvio Amir, Byron~C Wallace, Hao Lyu, and Paula Carvalho M{\'a}rio~J Silva.
\newblock Modelling context with user embeddings for sarcasm detection in
  social media.
\newblock \emph{arXiv preprint arXiv:1607.00976}, 2016.

\bibitem[Astudillo et~al.(2015)Astudillo, Amir, Ling, Silva, and
  Trancoso]{astudillo-EtAl:2015:ACL-IJCNLP}
Ram\'{o}n Astudillo, Silvio Amir, Wang Ling, Mario Silva, and Isabel Trancoso.
\newblock Learning word representations from scarce and noisy data with
  embedding subspaces.
\newblock In \emph{Proceedings of the 53rd Annual Meeting of the Association
  for Computational Linguistics and the 7th International Joint Conference on
  Natural Language Processing (Volume 1: Long Papers)}, pages 1074--1084,
  Beijing, China, July 2015.

\bibitem[Bamman and Smith(2015)]{bamman2015contextualized}
David Bamman and Noah~A Smith.
\newblock Contextualized sarcasm detection on twitter.
\newblock In \emph{Proceedings of the 9th International Conference on Web and
  Social Media}, pages 574--77. AAAI Menlo Park, CA, 2015.

\bibitem[Bengio et~al.(2003)Bengio, Ducharme, Vincent, and
  Janvin]{bengio2003neural}
Yoshua Bengio, R{\'e}jean Ducharme, Pascal Vincent, and Christian Janvin.
\newblock A neural probabilistic language model.
\newblock \emph{The Journal of Machine Learning Research}, 3:\penalty0
  1137--1155, 2003.

\bibitem[Bengio et~al.(2015{\natexlab{a}})Bengio, Goodfellow, and
  Courville]{Bengio-et-al-2015-Book}
Yoshua Bengio, Ian~J. Goodfellow, and Aaron Courville.
\newblock Deep learning.
\newblock Book in preparation for MIT Press, 2015{\natexlab{a}}.
\newblock URL \url{http://www.iro.umontreal.ca/~bengioy/dlbook}.

\bibitem[Bengio et~al.(2015{\natexlab{b}})Bengio, Goodfellow, and
  Courville]{bengio2015deep}
Yoshua Bengio, Ian~J Goodfellow, and Aaron Courville.
\newblock Deep learning.
\newblock \emph{Nature}, 521:\penalty0 436--444, 2015{\natexlab{b}}.

\bibitem[Blei et~al.(2003)Blei, Ng, and Jordan]{blei2003latent}
David~M Blei, Andrew~Y Ng, and Michael~I Jordan.
\newblock Latent dirichlet allocation.
\newblock \emph{the Journal of machine Learning research}, 3:\penalty0
  993--1022, 2003.

\bibitem[Brown et~al.(1992)Brown, Desouza, Mercer, Pietra, and
  Lai]{brown1992class}
Peter~F. Brown, Peter~V. Desouza, Robert~L. Mercer, Vincent J.~Della Pietra,
  and Jenifer~C. Lai.
\newblock Class-based n-gram models of natural language.
\newblock \emph{Computational Linguistics}, 18\penalty0 (4), 1992.

\bibitem[CDC(2010)]{centers2010current}
CDC.
\newblock Current depression among adults---united states, 2006 and 2008.
\newblock \emph{MMWR. Morbidity and mortality weekly report}, 59\penalty0
  (38):\penalty0 1229, 2010.

\bibitem[Collobert et~al.(2011)Collobert, Weston, Bottou, Karlen, Kavukcuoglu,
  and Kuksa]{collobert2011natural}
Ronan Collobert, Jason Weston, L{\'e}on Bottou, Michael Karlen, Koray
  Kavukcuoglu, and Pavel Kuksa.
\newblock Natural language processing (almost) from scratch.
\newblock \emph{The Journal of Machine Learning Research}, 12:\penalty0
  2493--2537, 2011.

\bibitem[Conway(2014)]{conway2014ethical}
Mike Conway.
\newblock Ethical issues in using twitter for public health surveillance and
  research: developing a taxonomy of ethical concepts from the research
  literature.
\newblock \emph{Journal of medical Internet research}, 16\penalty0
  (12):\penalty0 e290, 2014.

\bibitem[Coppersmith et~al.(2014{\natexlab{a}})Coppersmith, Dredze, and
  Harman]{coppersmith_acl_2014}
Glen Coppersmith, Mark Dredze, and Craig Harman.
\newblock Quantifying mental health signals in {T}witter.
\newblock In \emph{Proceedings of the ACL Workshop on Computational Linguistics
  and Clinical Psychology}, 2014{\natexlab{a}}.

\bibitem[Coppersmith et~al.(2014{\natexlab{b}})Coppersmith, Harman, and
  Dredze]{coppersmith_icwsm_2014}
Glen Coppersmith, Craig Harman, and Mark Dredze.
\newblock Measuring post traumatic stress disorder in {T}witter.
\newblock In \emph{Proceedings of the 8th International AAAI Conference on
  Weblogs and Social Media (ICWSM)}, 2014{\natexlab{b}}.

\bibitem[Coppersmith et~al.(2015{\natexlab{a}})Coppersmith, Dredze, Harman, and
  Hollingshead]{coppersmith_cross_condition_2015}
Glen Coppersmith, Mark Dredze, Craig Harman, and Kristy Hollingshead.
\newblock From {ADHD} to {SAD}: Analyzing the language of mental health on
  {T}witter through self-reported diagnoses.
\newblock In \emph{Proceedings of the Workshop on Computational Linguistics and
  Clinical Psychology: From Linguistic Signal to Clinical Reality}, Denver,
  Colorado, USA, June 2015{\natexlab{a}}. North American Chapter of the
  Association for Computational Linguistics.

\bibitem[Coppersmith et~al.(2015{\natexlab{b}})Coppersmith, Dredze, Harman,
  Hollingshead, and Mitchell]{shared_task_2015_overview}
Glen Coppersmith, Mark Dredze, Craig Harman, Kristy Hollingshead, and Margaret
  Mitchell.
\newblock {CLPsych} 2015 shared task: Depression and {PTSD} on {T}witter.
\newblock In \emph{Proceedings of the Shared Task for the NAACL Workshop on
  Computational Linguistics and Clinical Psychology}, 2015{\natexlab{b}}.

\bibitem[Coppersmith et~al.(2016)Coppersmith, Ngo, Leary, and
  Wood]{coppersmith_suicide_clpsych_2016}
Glen Coppersmith, Kim Ngo, Ryan Leary, and Tony Wood.
\newblock Exploratory data analysis of social media prior to a suicide attempt.
\newblock In \emph{Proceedings of the Workshop on Computational Linguistics and
  Clinical Psychology: From Linguistic Signal to Clinical Reality}, San Diego,
  California, USA, June 2016. North American Chapter of the Association for
  Computational Linguistics.

\bibitem[De~Choudhury et~al.(2016)De~Choudhury, Kiciman, Dredze, Coppersmith,
  and Kumar]{de2016discovering}
Munmun De~Choudhury, Emre Kiciman, Mark Dredze, Glen Coppersmith, and Mrinal
  Kumar.
\newblock Discovering shifts to suicidal ideation from mental health content in
  social media.
\newblock In \emph{Proceedings of the 2016 CHI Conference on Human Factors in
  Computing Systems}, pages 2098--2110. ACM, 2016.

\bibitem[Goldberg(2016)]{goldberg2016primer}
Yoav Goldberg.
\newblock A primer on neural network models for natural language processing.
\newblock \emph{Journal of Artificial Intelligence Research}, 57:\penalty0
  345--420, 2016.

\bibitem[Le and Mikolov(2014)]{le2014distributed}
Quoc~V Le and Tomas Mikolov.
\newblock Distributed representations of sentences and documents.
\newblock \emph{arXiv preprint arXiv:1405.4053}, 2014.

\bibitem[Li et~al.(2015)Li, Ritter, and Jurafsky]{li2015learning}
Jiwei Li, Alan Ritter, and Dan Jurafsky.
\newblock Learning multi-faceted representations of individuals from
  heterogeneous evidence using neural networks.
\newblock \emph{arXiv preprint arXiv:1510.05198}, 2015.

\bibitem[McCaughey et~al.(2014)McCaughey, Baumgardner, Gaudes, LaRochelle, Wu,
  and Raichura]{mccaughey2014best}
Deirdre McCaughey, Catherine Baumgardner, Andrew Gaudes, Dominique LaRochelle,
  Kayla~Jiaxin Wu, and Tejal Raichura.
\newblock Best practices in social media: Utilizing a value matrix to assess
  social media’s impact on health care.
\newblock \emph{Social Science Computer Review}, 32\penalty0 (5):\penalty0
  575--589, 2014.

\bibitem[Mikal et~al.(2016)Mikal, Hurst, and Conway]{mikal2016ethical}
Jude Mikal, Samantha Hurst, and Mike Conway.
\newblock Ethical issues in using twitter for population-level depression
  monitoring: a qualitative study.
\newblock \emph{BMC medical ethics}, 17\penalty0 (1):\penalty0 1, 2016.

\bibitem[Mikolov et~al.(2013)Mikolov, Sutskever, Chen, Corrado, and
  Dean]{mikolov2013distributed}
Tomas Mikolov, Ilya Sutskever, Kai Chen, Greg~S Corrado, and Jeff Dean.
\newblock Distributed representations of words and phrases and their
  compositionality.
\newblock In \emph{Advances in neural information processing systems}, pages
  3111--3119, 2013.

\bibitem[Mitchell et~al.(2015)Mitchell, Coppersmith, and
  Hollingshead]{clpsych_workshop_2015}
Margaret Mitchell, Glen Coppersmith, and Kristy Hollingshead, editors.
\newblock \emph{Proceedings of the Workshop on Computational Linguistics and
  Clinical Psychology: From Linguistic Signal to Clinical Reality}.
\newblock North American Association for Computational Linguistics, Denver,
  Colorado, USA, June 2015.

\bibitem[Paul and Dredze(2011)]{paul2011you}
Michael~J Paul and Mark Dredze.
\newblock You are what you tweet: Analyzing twitter for public health.
\newblock \emph{Icwsm}, 20:\penalty0 265--272, 2011.

\bibitem[Paykel et~al.(1997)Paykel, Tylee, Wright, Priest,
  et~al.]{paykel1997defeat}
ES~Paykel, A~Tylee, A~Wright, RG~Priest, et~al.
\newblock The defeat depression campaign: psychiatry in the public arena.
\newblock \emph{The American journal of psychiatry}, 154\penalty0 (6):\penalty0
  59, 1997.

\bibitem[Pedersen(2015)]{pedersen2015screening}
Ted Pedersen.
\newblock Screening twitter users for depression and ptsd with lexical decision
  lists.
\newblock In \emph{Proceedings of the Workshop on Computational Linguistics and
  Clinical Psychology: From Linguistic Signal to Clinical Reality}, pages
  46--53, 2015.

\bibitem[Pennebaker et~al.(2001)Pennebaker, Francis, and
  Booth]{pennebaker2001linguistic}
James~W Pennebaker, Martha~E Francis, and Roger~J Booth.
\newblock Linguistic inquiry and word count: Liwc 2001.
\newblock \emph{Mahway: Lawrence Erlbaum Associates}, 71:\penalty0 2001, 2001.

\bibitem[Preotiuc-Pietro et~al.(2015)Preotiuc-Pietro, Sap, Schwartz, and
  Ungar]{preotiuc2015mental}
Daniel Preotiuc-Pietro, Maarten Sap, H~Andrew Schwartz, and LH~Ungar.
\newblock Mental illness detection at the world well-being project for the
  clpsych 2015 shared task.
\newblock \emph{NAACL HLT 2015}, page~40, 2015.

\bibitem[Rajadesingan et~al.(2015)Rajadesingan, Zafarani, and
  Liu]{rajadesingan2015sarcasm}
Ashwin Rajadesingan, Reza Zafarani, and Huan Liu.
\newblock Sarcasm detection on twitter: A behavioral modeling approach.
\newblock In \emph{Proceedings of the Eighth ACM International Conference on
  Web Search and Data Mining}, pages 97--106. ACM, 2015.

\bibitem[{\v R}eh{\r u}{\v r}ek and Sojka(2010)]{rehurek_lrec}
Radim {\v R}eh{\r u}{\v r}ek and Petr Sojka.
\newblock {Software Framework for Topic Modelling with Large Corpora}.
\newblock In \emph{{Proceedings of the LREC 2010 Workshop on New Challenges for
  NLP Frameworks}}, pages 45--50, Valletta, Malta, May 2010. ELRA.
\newblock \url{http://is.muni.cz/publication/884893/en}.

\bibitem[Resnik et~al.(2015)Resnik, Armstrong, Claudino, Nguyen, Nguyen, and
  Boyd-Graber]{resnik2015beyond}
Philip Resnik, William Armstrong, Leonardo Claudino, Thang Nguyen, Viet-An
  Nguyen, and Jordan Boyd-Graber.
\newblock Beyond lda: exploring supervised topic modeling for
  depression-related language in twitter.
\newblock In \emph{Proceedings of the 2nd Workshop on Computational Linguistics
  and Clinical Psychology: From Linguistic Signal to Clinical Reality}, pages
  99--107, 2015.

\bibitem[Rumelhart et~al.(1988)Rumelhart, Hinton, and
  Williams]{rumelhart1988learning}
David~E Rumelhart, Geoffrey~E Hinton, and Ronald~J Williams.
\newblock Learning representations by back-propagating errors.
\newblock \emph{Cognitive modeling}, 5\penalty0 (3):\penalty0 1, 1988.

\bibitem[Schwartz et~al.(2014)Schwartz, Eichstaedt, Kern, Park, Sap, Stillwell,
  Kosinski, and Ungar]{schwartz2014towards}
H.~Andrew Schwartz, Johannes Eichstaedt, Margaret~L. Kern, Gregory Park,
  Maarten Sap, David Stillwell, Michal Kosinski, and Lyle Ungar.
\newblock Towards assessing changes in degree of depression through {Facebook}.
\newblock In \emph{Proceedings of the ACL Workshop on Computational Linguistics
  and Clinical Psychology}, 2014.

\bibitem[Severyn and Moschitti(2015)]{severyn2015twitter}
Aliaksei Severyn and Alessandro Moschitti.
\newblock Twitter sentiment analysis with deep convolutional neural networks.
\newblock In \emph{Proceedings of the 38th International ACM SIGIR Conference
  on Research and Development in Information Retrieval}, pages 959--962. ACM,
  2015.

\bibitem[Smith and Eisner(2005)]{smith2005contrastive}
Noah~A Smith and Jason Eisner.
\newblock Contrastive estimation: Training log-linear models on unlabeled data.
\newblock In \emph{Proceedings of the 43rd Annual Meeting on Association for
  Computational Linguistics}, pages 354--362. Association for Computational
  Linguistics, 2005.

\bibitem[Tang et~al.(2015)Tang, Qin, and Liu]{tang2015learning}
Duyu Tang, Bing Qin, and Ting Liu.
\newblock Learning semantic representations of users and products for document
  level sentiment classification.
\newblock In \emph{ACL (1)}, pages 1014--1023, 2015.

\bibitem[Van~der Maaten and Hinton(2008)]{van2008visualizing}
Laurens Van~der Maaten and Geoffrey Hinton.
\newblock Visualizing data using t-sne.
\newblock \emph{Journal of Machine Learning Research}, 9\penalty0
  (2579-2605):\penalty0 85, 2008.

\bibitem[Yang et~al.(2016)Yang, Chang, and Eisenstein]{yang2016toward}
Yi~Yang, Ming-Wei Chang, and Jacob Eisenstein.
\newblock Toward socially-infused information extraction: Embedding authors,
  mentions, and entities.
\newblock \emph{arXiv preprint arXiv:1609.08084}, 2016.

\bibitem[Yu et~al.(2016)Yu, Wan, and Zhou]{yu2016user}
Yang Yu, Xiaojun Wan, and Xinjie Zhou.
\newblock User embedding for scholarly microblog recommendation.
\newblock In \emph{Proceedings of the 54th Annual Meeting of the Association
  for Computational Linguistics}, volume~2, pages 449--453, 2016.

\end{thebibliography}

\newpage
\appendix
\section{Measuring Homophily (continued)}

\begin{figure}[h]
\subfloat[User vector similarity ranking. The first row corresponds to a 'query' user, and the columns show the top 100 most similar users, colored according to their class.]{%
  \includegraphics[width=1\columnwidth]{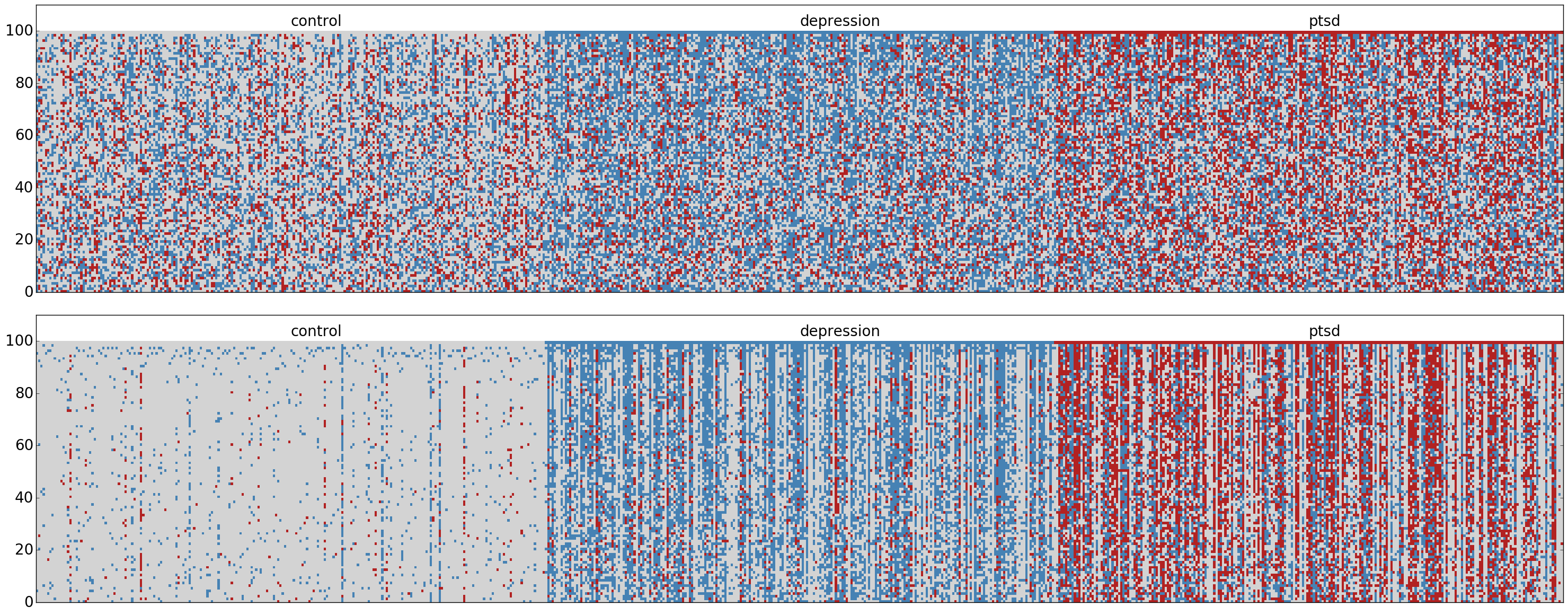}
  \label{fig:u2v_rank}
}
~\\
\subfloat[ROC curves and AUC scores of the induced user similarity rankings, per class.]{%
 \centering

    \includegraphics[width=1\columnwidth]{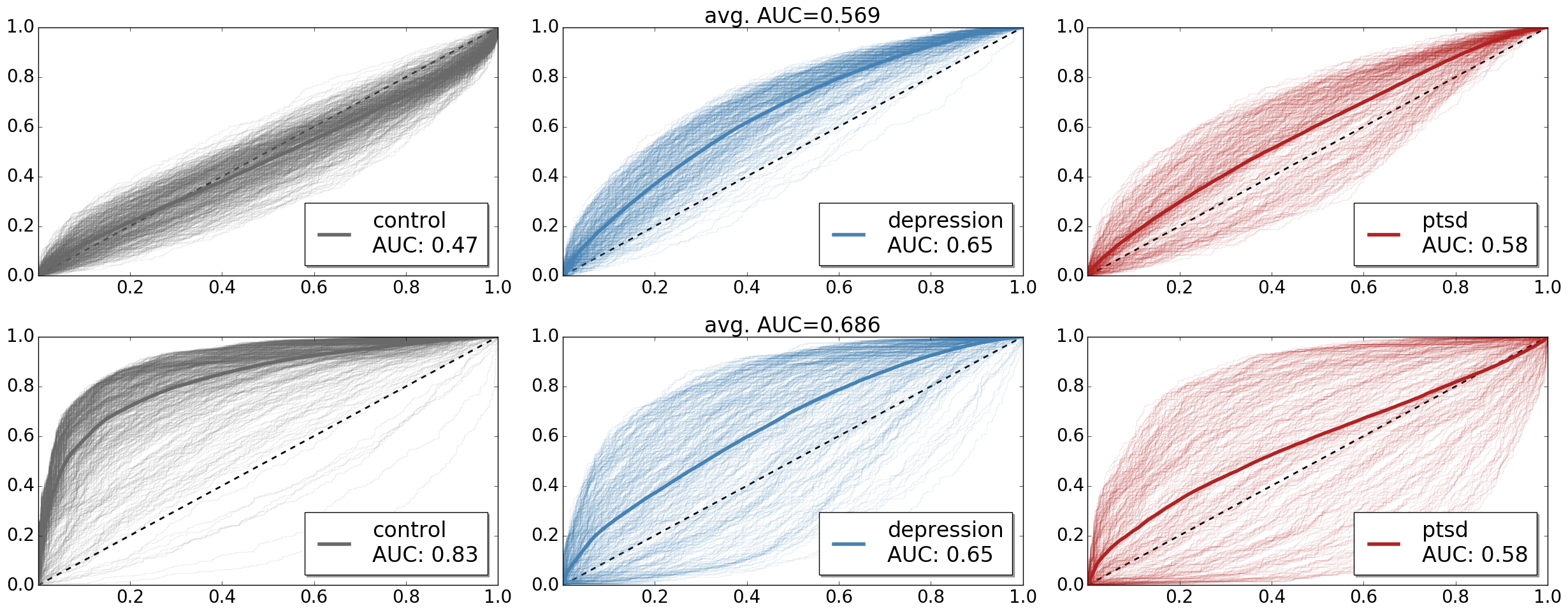}  
    \label{fig:u2v_roc}

}

\caption{Measuring homophilic relations with respect to mental conditions with vector distances over the user embedding space. The top-most sub-plots refer to rankings induced with the \textsc{User2Vec} model, and the ones at the bottom correspond to rankings obtained with embeddings adapted with the NLSE model.}
\label{fig:u2v_homophily}
\end{figure}

\begin{figure}[h]
\subfloat[User vector similarity ranking. The first row corresponds to a 'query' user, and the columns show the top 100 most similar users, colored according to their class.]{%
  \includegraphics[width=1\columnwidth]{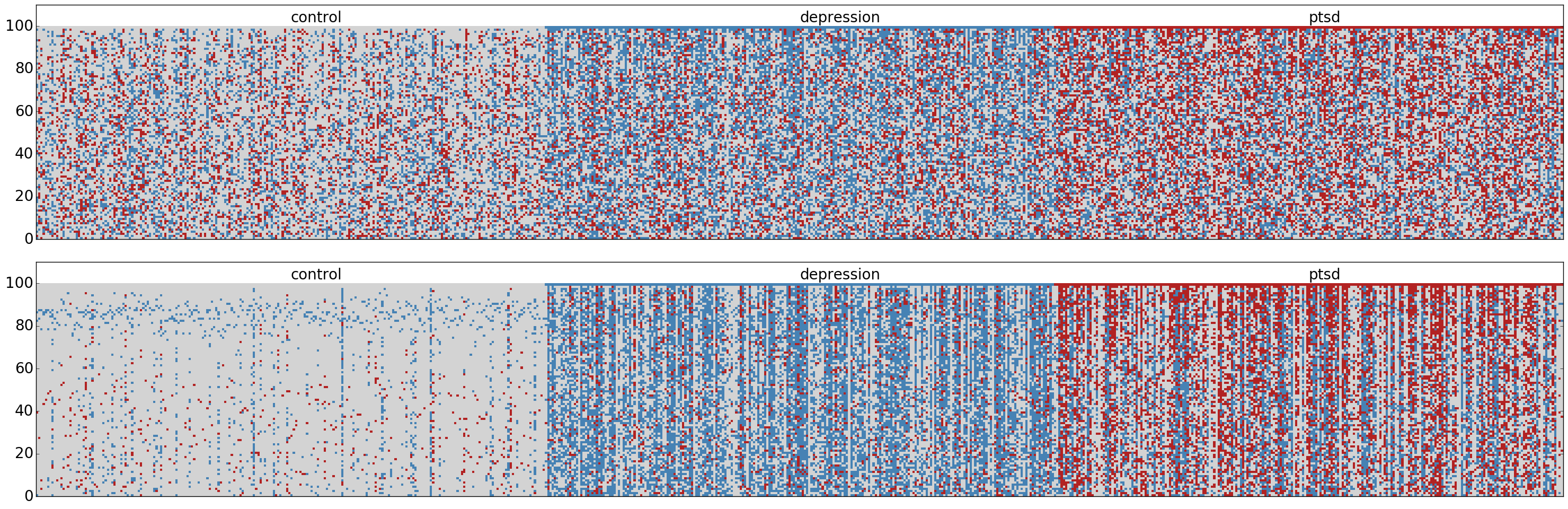}
  \label{fig:pvdbow_rank}
}
~\\
\subfloat[ROC curves and AUC scores of the induced user similarity rankings, per class.]{%
 \centering

    \includegraphics[width=1\columnwidth]{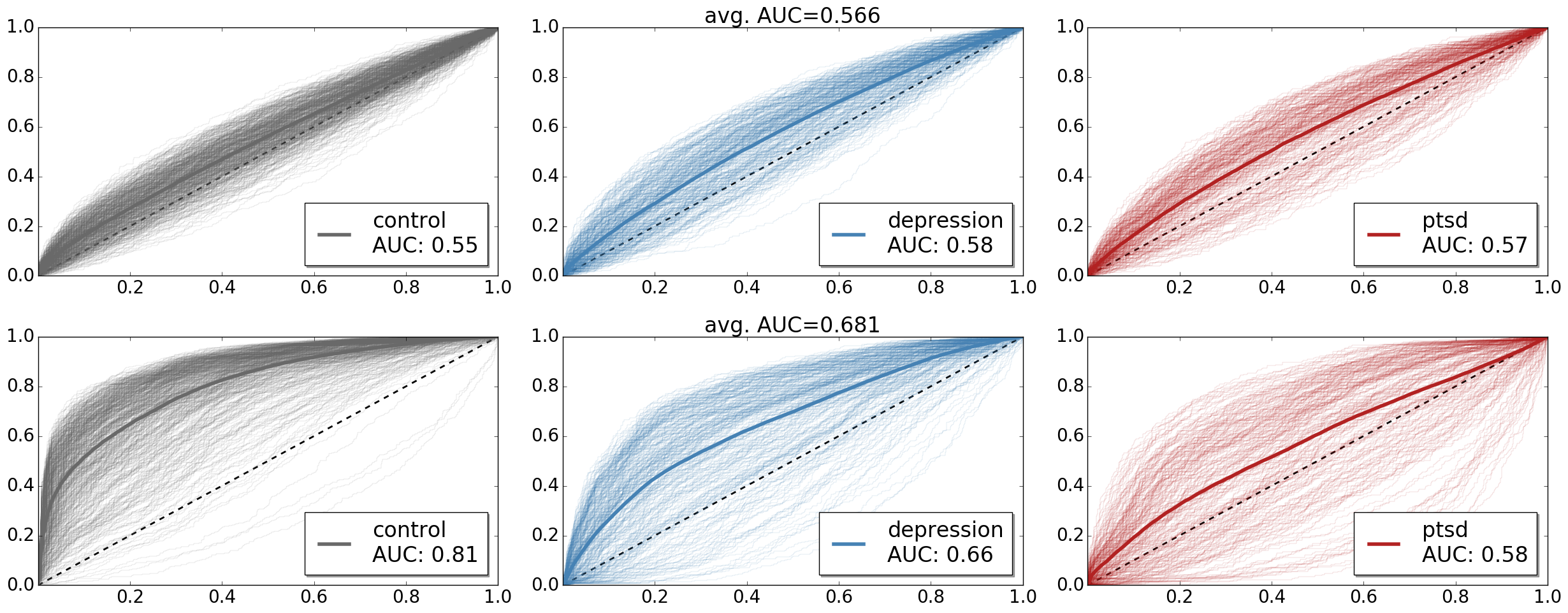}  
    \label{fig:pvdbow_roc}

}

\caption{Measuring homophilic relations with respect to mental conditions with vector distances over the user embedding space. The top-most sub-plots refer to rankings induced with the \textsc{PV-dbow} model, and the ones at the bottom correspond to rankings obtained with embeddings adapted with the NLSE model.}
\label{fig:pvdbow_homophily}
\end{figure}

\section{Visualizing Embedding Subspace Features}

\begin{figure}[h]
\subfloat{%
  \includegraphics[width=1\columnwidth]{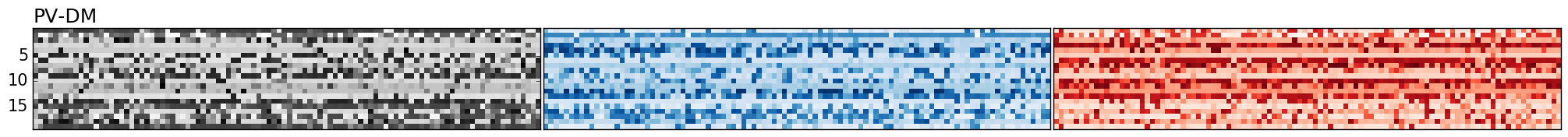}  
  \label{fig:sub_pvdm}
}
~\\
\subfloat{%
 \centering
    \includegraphics[width=1\columnwidth]{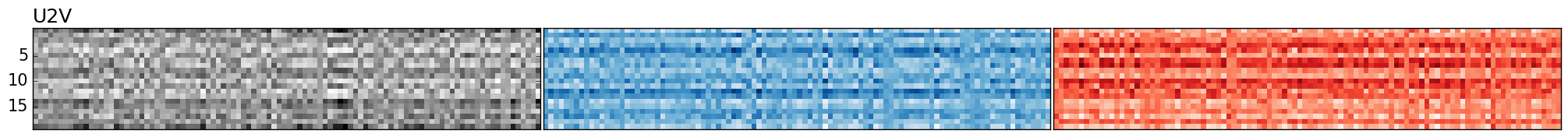}
    \label{fig:sub_u2v}
}
~\\
\subfloat{%
 \centering
    \includegraphics[width=1\columnwidth]{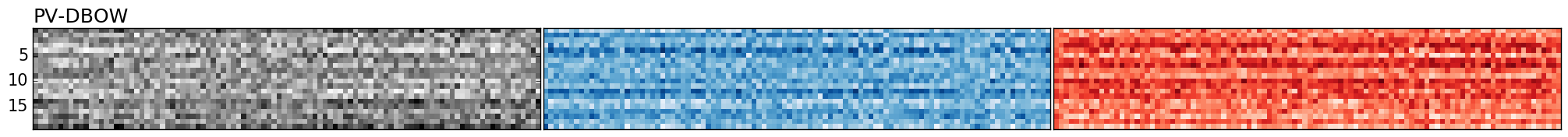}
    \label{fig:sub_pvdbow}
}
\caption{Embedding subspace features by class. Color intensities reflect the magnitude of the feature values}
\label{fig:subspace_feats}
\end{figure}

\begin{figure}[h]
\centering
\subfloat{%
  \includegraphics[width=0.6\columnwidth]{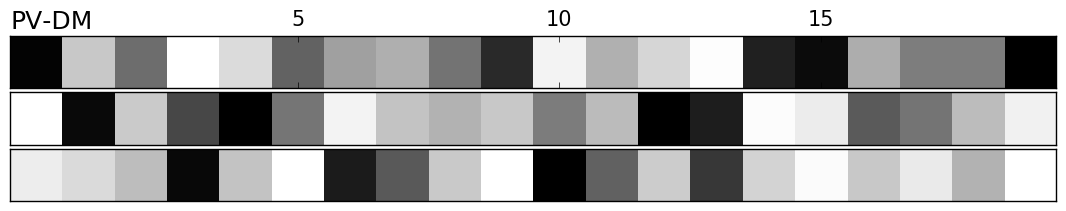}
  \label{fig:sub_pvdm_mean}
}
~\\
\subfloat{%
 \centering
    \includegraphics[width=0.6\columnwidth]{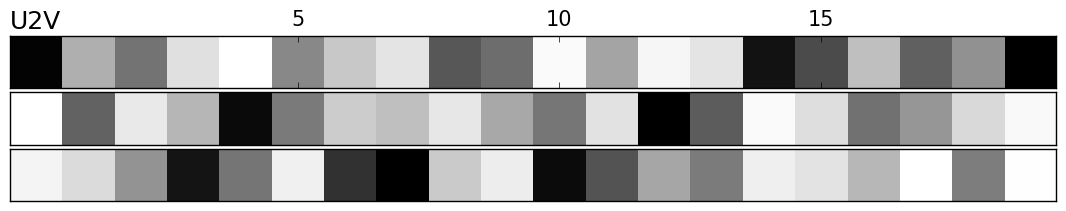}
    \label{fig:sub_u2v_mean}
}
~\\
\subfloat{%
 \centering
    \includegraphics[width=0.6\columnwidth]{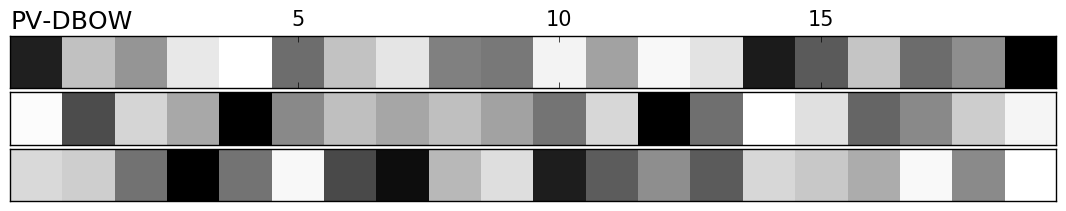}
    \label{fig:sub_pvdbow_mean}
}
\caption{Embedding subspace features for a `prototypical class vector' obtained by averaging the vectors of all the users in the  class.  }
\label{fig:subspace_feats_mean}
\end{figure}

The NLSE induces task-specific representations by learning a low dimensional embedding. We exploit the fact that the sigmoid non-linearity (Eq. \ref{eq:nlse}) projects all the feature values into the range $[0;1]$, to map these values into color intensities in a heatmap. We then used the same sample used to produce the plots in Section \ref{sec:embedding_analysis} (100 users per class), and plotted the representations learnt by model. The resulting plot is shown in Figure \ref{fig:subspace_feats}. We can observe that specific groups of features are mostly activated in specific classes. From this plot, we can see that the features induced with \textsc{PV-DM} model are sparser than the other models, which might explain why these features can better capture user similarities. On the other hand, the features induced with \textsc{PV-DM} and \textsc{User2Vec} seem `noisier' but also seem to capture differences between classes. In Figure \ref{fig:subspace_feats_mean}, we show a similar plot but where we average the feature vectors of \emph{all} the users in each class. Interestingly, it seems to be case that the `prototyipcal' class vectors learned with different embeddings are very similar.

\end{document}